\documentclass[journal]{IEEEtran}
\usepackage[caption=false]{subfig}

\usepackage{xcolor,soul,framed} 

\usepackage[pdftex]{graphicx}
\graphicspath{{../pdf/}{../jpeg/}}
\DeclareGraphicsExtensions{.pdf,.jpeg,.png}

\usepackage[cmex10]{amsmath}
\usepackage{array}
\usepackage{mdwmath}
\usepackage{mdwtab}
\usepackage{eqparbox}
\usepackage{url}
\usepackage{multirow}
\usepackage{booktabs}
\usepackage[compatibility=false]{caption}
\usepackage{graphicx}
\usepackage{diagbox}
\usepackage{amsmath}
\usepackage{float}
\usepackage{booktabs}
\usepackage{booktabs,makecell}
\usepackage{siunitx}
\usepackage{pgfplots}
\pgfplotsset{compat=1.18}
\usepackage{pgfplotstable}
\usepackage{caption}
\usepackage{siunitx}
\renewcommand{\arraystretch}{1.2}
\usepackage{diagbox}
\usepackage{makecell}
\graphicspath{{figures/}}

\begin{document}
\bstctlcite{IEEEexample:BSTcontrol}
    \title{Enabling Federated Object Detection for Connected Autonomous Vehicles: A Deployment-Oriented Evaluation}
  \author{Komala Subramanyam~Cherukuri,~\IEEEmembership{Student Member,~IEEE,}
      Kewei~Sha,~\IEEEmembership{Senior Member,~IEEE,} Zhenhua~Huang} 



\maketitle

\begin{abstract}
Object detection is crucial for Connected Autonomous Vehicles (CAVs) to perceive their surroundings and make safe driving decisions. Centralized training of object detection models often achieves promising accuracy, fast convergence, and simplified training process, but it falls short in scalability, adaptability, and privacy-preservation. Federated learning (FL), by contrast, enables collaborative, privacy-preserving, and continuous training across naturally distributed CAV fleets. However, deploying FL in real-world CAVs remains challenging due to the substantial computational demands of training and inference, coupled with highly diverse operating conditions. Practical deployment must address three critical factors: (i) heterogeneity from non-IID data distributions, (ii) constrained onboard computing hardware, and (iii) environmental variability such as lighting and weather, alongside systematic evaluation to ensure reliable performance. This work introduces the first holistic deployment-oriented evaluation of FL-based object detection in CAVs, integrating model performance, system-level resource profiling, and environmental robustness. Using state-of-the-art detectors, YOLOv5, YOLOv8, YOLOv11, and Deformable DETR, evaluated on the KITTI, BDD100K, and nuScenes datasets, we analyze trade-offs between detection accuracy, computational cost, and resource usage under diverse resolutions, batch sizes, weather and lighting conditions, and dynamic client participation, paving the way for robust FL deployment in CAVs. 

\end{abstract}


%
\IEEEpeerreviewmaketitle


\section{\textbf{Introduction}}

Connected Autonomous Vehicles (CAVs) are revolutionizing intelligent transportation systems by integrating core tasks such as perception, prediction, planning, control, V2X communication, and system level functions \cite{yurtsever2020survey, zhao2025survey}. Perception is the most critical component of this pipeline because every downstream function 
is dependent on a precise understanding of the environment. Within perception, object detection is the core task, as it enables vehicles to identify, classify, and localize surrounding vehicles, pedestrians, and obstacles in real time \cite{sun2023toward}. CAVs achieve this capability by combining data from heterogeneous sensors such as LiDAR, cameras, and radar to create a comprehensive and reliable representation of the driving environment, which is essential for safe and intelligent autonomous operation \cite{jebamikyous2022autonomous}.

Traditional deep learning for CAV perception relies on centralized learning, where large volumes of data are transferred from distributed vehicles to a central server for model training. This approach results in high communication overhead, exposes sensitive data to privacy risks, and is fundamentally incompatible with the decentralized and heterogeneous nature of vehicular networks  \cite{nguyen2021federated}. Federated learning (FL) overcomes these limitations by allowing vehicles and road units to train models locally without sharing raw data \cite{fu2024secure}. By reducing communication costs, preserving privacy, and supporting heterogeneous non-i.i.d. and device environments, FL aligns naturally with the requirements of large-scale CAV deployment \cite{chellapandi2023federated, zhang2024survey}.

Even though promising, deploying FL in CAVs requires each vehicle to process sufficient computing power, storage, and communication capabilities onboard \cite{imteaj2021survey}, especially for real-time object detections based on large scale (e.g., one gigabyte of sensor data per second \cite{elbir2022federated}) multi-modal data from various sensors such as camera, LiDAR, and radar. Beyond these requirements, FL in CAVs faces significant challenges arising from heterogeneity \cite{elbir2022federated} across data distributions, sensor configurations, hardware capabilities, communication conditions, and power budgets \cite{moon2024client}, all of which vary substantially from vehicle to vehicle. Such disparities frequently result in inconsistent participation, client dropouts, and reduced scalability during federated training \cite{lu2024federated}. In addition, environmental conditions introduce further variability that lies beyond system control, compounding the difficulty of achieving robust and generalizable models. Moreover, Object detection requires real-time processing of high-resolution sensor data and thousands of bounding-box predictions per frame, imposing heavy memory, compute, and energy demands, and must not interfere with other safety-critical modules \cite{khan2024real, wu2017squeezedet}.

While these challenges highlight the promise and complexity of FL in CAVs, they also expose a fundamental gap between algorithmic development and real-world deployment. Existing FL studies evaluate performance primarily through accuracy or convergence, but these metrics overlook critical system-level constraints such as latency, resource usage, and client availability that determine feasibility. As a result, models that appear to be effective in experiments may still be impractical or unsafe in deployment. Therefore, a comprehensive evaluation must account for computational efficiency, memory footprint, energy consumption, and robustness under heterogeneous conditions to ensure that FL frameworks are both scalable and reliable in real-world CAV systems.

Evaluating federated object detection models for CAVs is critical to ensuring real world readiness and safe deployment. Such evaluation must address three key dimensions:

\begin{itemize}
\item \textbf{Data Heterogeneity:} Evaluation assesses the impact of non-IID data across CAVs to understand convergence behavior and guide strategies for model generalization.  
\item \textbf{System Resources:} Evaluation identifies whether models remain feasible under real-time limits of computation, memory, energy, and bandwidth across diverse hardware. 
\item \textbf{Client and Environment Diversity:} Evaluation assesses model robustness under varying client participation, and environmental conditions such as lighting and weather. 
\end{itemize}

A comprehensive evaluation is essential to balance accuracy, efficiency and safety in FL-based perception, taking into account variations in hardware, sensors, and data distributions to ensure a reliable and scalable deployment. Such evaluation is necessary to ensure that models can be integrated without disrupting other safety-critical tasks in CAVs. However, testing these evaluations directly in real CAVs is constrained by safety concerns, regulatory restrictions, and prohibitive costs. Simulation-based frameworks therefore become essential to study scalability, resource behavior, and deployment readiness under controlled yet realistic conditions.

In this work, we develop an evaluation framework built on the Flower platform to address these challenges. We adopted Flower framework because of its modularity, and ability to simulate large federated environments at low cost, making it well-suited for experimentation under deployment conditions. To extend its utility for autonomous driving, we incorporate system-level profiling to capture critical metrics such as memory usage, power consumption, and training time, which are often overlooked in FL studies but are essential for the feasibility of deployment. Therefore, evaluation is the mechanism to translate experimental performance into deployment feasibility. Within this framework, we integrate state-of-the-art object detectors, YOLOv5, YOLOv8, YOLOv11, and Deformable DETR to evaluate their performance on the KITTI, BDD100K, and nuScenes, datasets. We further simulate deployment challenges including client variability, image size variability, and environmental inconsistency (e.g., lighting and weather) to assess model robustness and system behavior under non-ideal conditions.


The contributions of this paper are three-fold. First, we design a comprehensive FL evaluation framework that integrates the Flower platform with system-level resource profiling and dynamic client management, enabling realistic assessments under heterogeneous conditions. Second, we conduct a systematic performance evaluation using three widely adopted datasets, multiple state-of-the-art object-detection models, and representative aggregation algorithms under diverse experimental settings, thereby revealing key factors that directly affect the real-world deployment of FL in CAVs. Third, drawing on insights from this evaluation, we identify and articulate a set of open research problems that are critical to advancing the practical deployment of FL in CAV ecosystems. Collectively, these contributions provide both a foundation for rigorous benchmarking and a roadmap for future research in this domain.

The remainder of this paper is organized as follows. Sections~\ref{sec:Background} and Section~\ref{sec:motivation} present the background and motivation, respectively. Section~\ref{sec:framework} details the design of our evaluation framework, while Section~\ref{sec:experiment_setup} describes the datasets and metrics. Section~\ref{sec:results} presents the results and Section~\ref{sec:evaluation} discusses the limitations and open research problems. Finally, we conclude the paper in Section~\ref{sec:conclusion}.

\section{\textbf{Background}}
\label{sec:Background}

\subsection{\textbf{Object Detection Models}}

Object detection is a fundamental perception task in autonomous driving, responsible for both classifying and localizing surrounding objects using sensor modalities such as cameras, LiDAR, and radar \cite{feng2021review, li2020deep}. Convolutional neural networks (CNNs) have become the backbone of real-time detection pipelines, being one of the most widely adopted architectures for real-time applications \cite{turay2022toward, parambil2024navigating}. More recently, transformer-based architectures, such as Deformable DETR have been introduced to enhance detection in dense or complex scenes while achieving faster convergence compared to the original DETR \cite{zhu2020deformable}. These advances form the foundation for modern perception systems in CAVs and serve as the basis for further adaptation in FL.

\subsection{\textbf{Federated Learning in Connected Autonomous Vehicles}}

Autonomous vehicles can produce 20–40 TB of sensor data each day, largely from high-throughput perception systems such as cameras (20–60 Mbps), LiDAR (10–70 Mbps), sonar (10–100 kbps), and radar (10 kbps) \cite{wang2020hydraspace}. Transmitting this volume continuously is infeasible; for example, 5G mmWave may deliver only about 4 Mbps per vehicle \cite{khan2018feasibility}. These limitations make it unfeasible to transfer all raw sensor data to centralized servers for model training \cite{kim2022data}, which may lead to the loss of important information. In addition, CAVs require fast on-vehicle processing to ensure timely decision making.

\begin{figure}[H]
    \centering   \includegraphics[width=0.38\textwidth]{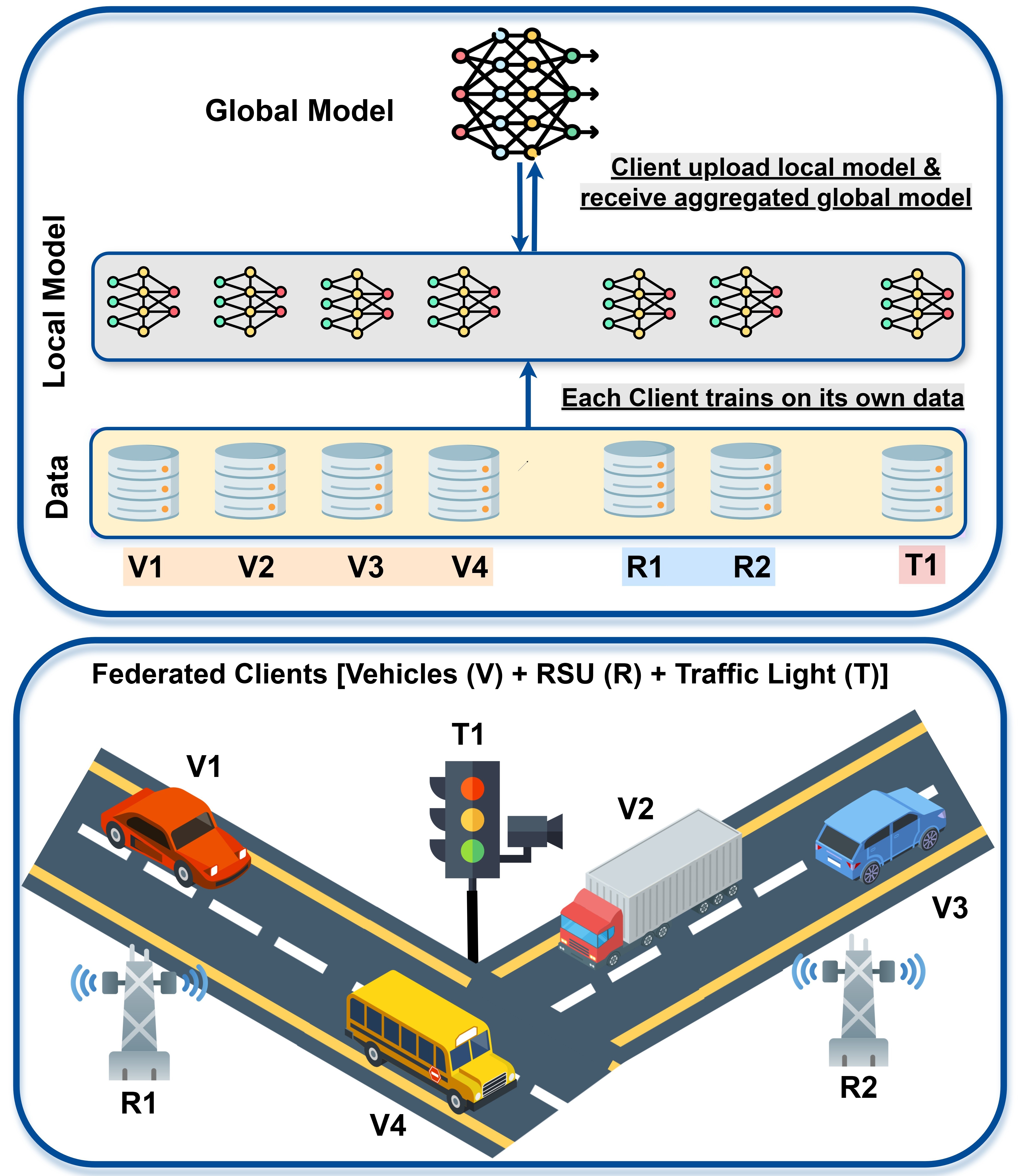}
    \caption{Federated Learning for CAVs}
    \label{fig:FLforCAV}
    \vspace{-0.2cm}
\end{figure}


As shown in Fig.~\ref{fig:FLforCAV} FL offers one such approach by enabling decentralized model training in which vehicles retain raw sensor data locally and share only model updates with a central server \cite{nakanoya2021personalized, li2021privacy}. This approach reduces communication overhead, preserves data privacy, and supports client-specific learning while avoiding the transmission of massive sensor streams \cite{sha2025privacy, fu2023incentive}. FL methods are categorized as:
\begin{itemize}
    \item Synchronous FL - The server aggregates updates after receiving them from all or most clients, improving accuracy but requiring reliable communication \cite{zhou2024adaptive}. 
    \item Asynchronous FL - The server aggregates updates as they arrive, enabling faster training and flexible participation, but potentially introducing model inconsistency due to varied update timing \cite{you2023federated}. 
\end{itemize}

These synchronous and asynchronous approaches have guided research on federated object detection for CAVs. Based on these principles, studies have explored various strategies to address performance, communication, and deployment challenges in real-world settings.

Architectural efficiency and perception enhancement have been major goals in FL for CAVs. Hierarchical FL methods combine synchronous and asynchronous updates for lightweight edge detection \cite{behera2024large}, while parameter-efficient designs for 3D object detection use adapter-based updates and roadside units to reduce communication costs \cite{chi2025parameter}. Some approaches adopt a sparse aggregation, where only a subset of model parameters or gradients is transmitted to reduce bandwidth usage in the Internet of Vehicles settings \cite{qian2024toward}. Federated transfer learning has also been applied to pedestrian detection, enabling adaptation under diverse weather and data-scarce conditions \cite{fittipaldi2024vehicular}.

Addressing the challenges of non-IID data distribution and heterogeneous client capabilities has been another priority in recent work. Semi-supervised FL improves the detection of static and dynamic objects by taking advantage of limited labeled data with larger pools of unlabeled samples \cite{chi2023federated}. Label aggregation combined with proximal regularization mitigates performance drops caused by heterogeneity of the data \cite{khalil2024federated}. Real-time deployments have been demonstrated using MPI-based FL for YOLOv7 in encrypted, heterogeneous environments \cite{quemeneur2024fedpylot}, while replay of experience in self-supervised depth estimation reduces catastrophic forgetting over time \cite{soares2024towards}.

Beyond model and data considerations, security is a critical requirement in FL for CAVs. Secure aggregation protocols with YOLOv3 protect model updates while maintaining accuracy comparable to centralized training \cite{jallepalli2021federated}. Interpretability has been enhanced using YOLOv7 augmented with Grad-CAM and CBAM, improving transparency in detection \cite{devarajan2025explainable}.

To handle heterogeneous sensors and diverse perception inputs, multimodality and cooperative perception address the challenges posed by heterogeneous sensors and diverse data modalities. Solutions include adaptation to loss functions, missing data imputation, and intelligent client selection \cite{zheng2023autofed}. Feature-level fusion and clustering improve 3D object detection \cite{chi2023fede}, while Bird’s Eye View methods incorporate camera-aware personalization \cite{song2023fedbevt} and dynamic aggregation with adaptive loss functions \cite{zhang2024federated}.
At the system level, FL frameworks have been developed to jointly optimize network resource allocation and sensor deployment \cite{wang2022federated}, along with hybrid privacy-preserving schemes that defend against data poisoning and inference attacks \cite{mia2024secure}.

\section{\textbf{Motivation}}
\label{sec:motivation}

Existing research has made progress in federated object detection for CAVs, yet real-world operation presents complex and interdependent constraints that are often overlooked and must be addressed to ensure reliable deployment. For reliable deployment, several inherent challenges must be addressed: (i) heterogeneity arising from non-IID data distributions, (ii) limitations of onboard computing hardware, and (iii) environmental variability such as lighting and weather. These factors significantly affect system performance, making systematic evaluation essential. Such evaluation not only ensures robustness under diverse conditions, but also provides critical insights into the feasibility of real-world deployment. We emphasize three key challenges: data heterogeneity, hardware constraints such as power, GPU memory, and utilization, and environmental conditions beyond human control.
\vspace{-0.25cm}
\subsection{\textbf{Heterogeneity}}
\begin{figure}[H]
    \centering
    \includegraphics[width=0.45\textwidth]{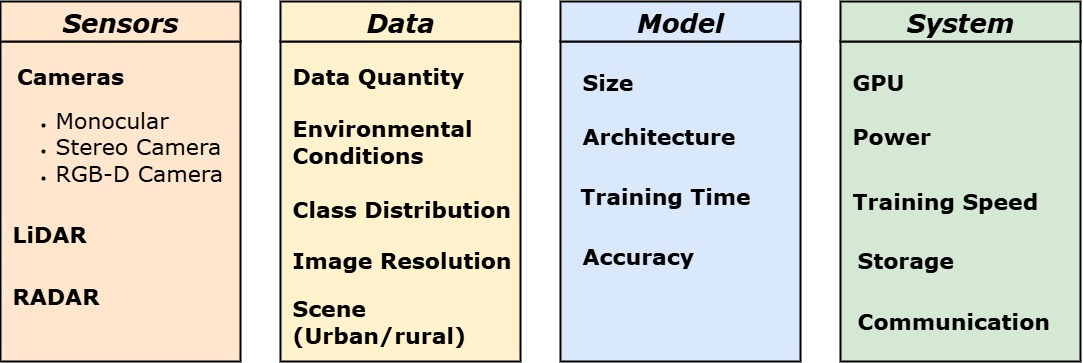}
    \caption{Heterogeneity in CAVs}
    \label{fig:HeteroinCAV}
    \vspace{-0.25cm}
\end{figure}

CAVs operate in highly dynamic environments, introducing substantial heterogeneity into FL systems. This heterogeneity arises from multiple sources, including differences in the size and resolution of perception data \cite{fang2024cooperative}, non-IID data distributions \cite{yin2023nipd}, variations in computational capabilities \cite{wu2024enhancing}, and fluctuating network conditions \cite{donevski2021addressing}. Beyond data and system-level disparities, CAVs also differ in their control dynamics; while vehicles may share a common model architecture, their parameters can vary due to factors such as acceleration response times or control gains \cite{hu2022hierarchical}. As illustrated in Fig.~\ref{fig:HeteroinCAV}, heterogeneity spans sensor types, the quantity and distribution of collected data \cite{zheng2023autofed, zheng2015reliable}, model architectures \cite{diao2020heterofl}, and system-level resources \cite{wu2022fedadapt}. In federated object detection, such heterogeneity can lead to inconsistent model updates, slower convergence, and degraded performance when aggregating knowledge across clients. Evaluating FL under heterogeneous client conditions is essential for understanding the practical feasibility of FL in CAV deployments.

\subsection{\textbf{System Constraints in Real World CAV Deployment}}
In addition to heterogeneity in data, models, and resources, CAV deployments face strict system-level constraints \cite{feng2020deep}. Differences in the sensor and processor configurations on board lead to wide variation in available computing resources \cite{xiong2020communication}. These limitations are compounded by the need to run multiple concurrent tasks such as perception, localization, planning, and control, each requiring dedicated models and competing for the same hardware. Two critical constraints in this regard are GPU memory and power consumption.

\subsubsection{\textbf{GPU Memory}}
Graphics Processing Units (GPUs) are essential for accelerating deep learning tasks such as object detection, as they can process large-scale parallel computations with high arithmetic intensity \cite{castano2008performance}. In CAVs, object detection requires both classification and precise localization of surrounding objects, making it computationally intensive and highly dependent on GPU acceleration \cite{efthymiadis2024advanced}. Modern GPUs with CUDA cores and specialized tensor units enable high-throughput training and inference for both centralized development and real-time embedded execution \cite{shin2022deep, jeon2021deep}. However, GPU memory capacity varies across FL clients for CAVs, affecting feasible batch sizes, image resolutions, and model complexity. Evaluating memory consumption is therefore critical for understanding its impact on scalability, reliability, and fairness in resource-constrained CAVs.

\subsubsection{\textbf{Power}}
CAVs rely on multiple sensors, processors, and communication modules, all contributing to high power demands. Embedded high-performance computing platforms can draw thousands of watts \cite{tang2020lopecs}, largely due to the computational intensity of models for tasks such as object detection. These models, with millions of parameters, require continuous real-time inference, which directly impacts energy efficiency and driving range \cite{lin2018architectural}. At scale, the cumulative energy demand onboard can increase greenhouse gas emissions \cite{kwak2024deep}. High-accuracy object detection alone can consume hundreds of watts \cite{xiong2024reducing}, making energy efficiency a critical consideration for the deployment. In FL-enabled CAVs, evaluating power consumption in realistic operational settings is essential to understand its implications for sustainable real-time operation.

\vspace{-0.3cm}

\subsection{\textbf{Environmental Complexity}}
The sensors, processors, and communication systems can be configured and maintained on board, environmental conditions remain outside human control. CAVs are highly dependent on sensor-based perception to navigate safely, but adverse weather and lighting changes can severely degrade sensor performance and alter the appearance of the surroundings \cite{tahir2024object}. Real-world evaluations of Level 2 autonomous systems clearly show these limitations. For example, Tesla Autopilot and GM Super Cruise have demonstrated reduced reliability in rain, snow, or poor visibility, at times causing over-steering or completely blocking autonomous functions \cite{Lambert2019,supercruise2018}. Although extreme conditions such as dense fog occur infrequently, accounting for only about 0.01\% of driving time in North America \cite{van2010roles}, their impact on perception accuracy is substantial. Weather effects such as rain, fog, and snow weaken sensor signals, introduce noise, and reduce data quality, which in turn affects perception performance and other tasks \cite{bijelic2020seeing,zhang2023perception}. These uncontrollable variations create additional data heterogeneity, as vehicles exposed to different environmental scenarios may develop inconsistent or even conflicting local models, making collaborative learning in federated systems more difficult. Evaluating how such environmental complexity impacts federated object detection is therefore essential to understanding real-world deployment performance.


To bridge the gap between FL research and its practical deployment in CAVs, we present a deployment-focused evaluation framework that systematically examines the object detection models perform under operationally realistic conditions. 

In this work, we selected the Flower framework due to its ability to simulate large-scale federated environments while offering flexibility to adapt and control over system behavior. We focused our evaluation on GPU memory usage, power consumption, and training time, as these represent the most critical constraints faced by real-world vehicle hardware during perception tasks. We chose YOLOv5, YOLOv8, YOLOv11, and Deformable DETR models because they span a wide spectrum of architectural complexity, resource requirements, and real-time applicability. Rather than modifying model architectures or training algorithms, we prioritized evaluating them in controlled scenarios that reflect real-world deployment challenges such as client dropout, variable image sizes, and adverse weather or lighting. This focused evaluation strategy was essential to isolate the key trade-offs between detection performance and resource usage. By focusing the study on rigorous evaluation instead of algorithmic modification, our work offers a novel, deployment-oriented perspective that bridges the gap between experimental FL research and the practical requirements of CAVs.

\section{\textbf{Evaluation Framework Design for Federated Object Detection in CAV}}
\label{sec:framework}

\begin{figure}[H]
    \centering
    \includegraphics[width=0.4\textwidth]{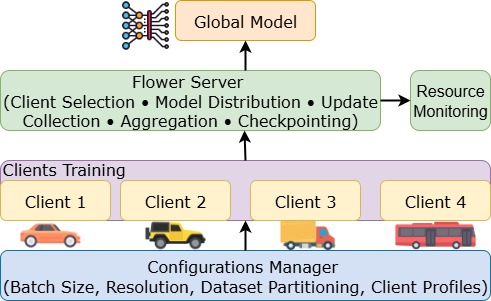}
    \caption{Evaluation Framework Design}
    \label{fig:setup}
    \vspace{-0.3cm}
\end{figure}

To address gaps in existing research on federated object detection, we present a comprehensive evaluation design built on the Flower framework as shown in Fig~\ref{fig:setup}. It supports privacy-preserving training of YOLOv5, YOLOv8, YOLOv11, and Deformable DETR in simulated CAV clients with non-IID data. Each client represents a CAV operating under varying computational constraints and data characteristics. Our task is conducted on Nvidia H100 NVL GPU. This powerful hardware configuration accelerates the processing and analysis of data, enhancing the speed.

The Flower framework provides a flexible foundation for orchestrating FL, it lacks built-in support for system level resource monitoring. To bridge this limitation, \textit{we design an evaluation framework that enables systematic assessment of resource consumption and model performance under deployment-oriented conditions}. Our framework facilitates systematic evaluation of object detection models under realistic operational constraints. We integrate psutil, pynvml, and nvidia-smi to track resource usage in real time. The system includes a model checkpoint for fault tolerance and supports adaptive participation to reflect real-world deployment.

To support both YOLO and Deformable DETR within a unified FL framework, several architectural and system-level considerations are required. Although the core FL logic, such as client-server communication, aggregation, and training flow, remains unchanged, the underlying models differ significantly in structure. YOLO is based on convolutional layers and is typically pre-trained for fast, while Deformable DETR relies on transformer-based attention mechanisms and operates with a different training pipeline and output format.

We standardized model initialization, parameter exchange, and output handling so that both detectors could run in the same FL environment without modifying the aggregation logic. YOLO pre-trained weights were reinitialized, so all clients began federated training from a common state. GPU memory profiling required careful handling because PyNVML reports device-level usage rather than per-client usage during sequential execution, so we log usage only during the active client training window to attribute it correctly. Next, we detail our unique designs customized for the identified evaluation goals by creating a simulated FL environment that resembles real-world CAV systems. 

\vspace{-0.34cm}
\subsection{\textbf{Federated Training Coordination using Flower}}
The Flower framework is used for custom client behavior, dynamic scheduling, and flexible integration. Each client simulates a CAV and performs local training on a partitioned, non-IID subset of the dataset. Flower handles model initialization, client selection, and aggregation of locally trained parameters at the end of each round. Client execution is coordinated using Ray for dynamic instantiation and runtime selection of clients based on resource availability. After local training, clients return updated weights to the server, which aggregates them using FedAvg or FedProx and redistributes the global model. This coordination ensures that all models are evaluated under identical FL dynamics. 
\vspace{-0.5cm}

\subsection{\textbf{Client Coordination and Model Management}}
Flower manages communication between server and clients each round. Clients train locally and return updates that the server aggregates. To prevent shape mismatches, all clients load an identical predefined checkpoint. This ensures that differences in results stem from model and data variations, not initialization inconsistencies.
\vspace{-0.5cm}

\subsection{\textbf{Model Flexibility and Federated Adaptability}}
Our framework support both lightweight YOLO and transformer-based Deformable DETR without major architectural changes. This is achieved by modularizing the model loading and training routines so that both architectures share the same aggregation logic and parameter exchange format. Evaluating both under identical FL conditions isolates architectural effects from orchestration effects, allowing fair accuracy and efficiency comparisons.
\vspace{-0.5cm}

\subsection{\textbf{System Resource Profiling}}
Resource constraints are critical for CAV deployment. We integrate pynvml to monitor GPU memory, power, utilization, and temperature, and psutil to track CPU load and RAM usage. The metrics are recorded before, during and after each training stage to evaluate the resource consumption patterns between different models, image resolutions, and batch sizes. This evaluation reveals configurations that achieve the accuracy of the target within the hardware limits, guiding the model hardware match for deployment.
\vspace{-0.5cm}

\subsection{\textbf{Controlled Experimental Conditions}}
Clients are executed sequentially on a single GPU to avoid memory contention and match evaluation in resource-limited CAVs. This isolation ensures stability and allows for accurate attribution of GPU usage to the active client. To ensure reproducibility, we enforce strict controls, a single pretrained checkpoint for all clients, a clean CUDA state before each run, and fixed random seeds. GPU memory is recorded immediately after local training and at intra-epoch checkpoints, validated with PyNVML queries. These measures ensure fair and repeatable comparisons of model performance and resource usage.
\vspace{-0.4cm}

\subsection{\textbf{Support for Variable Client Configurations}}
We simulate heterogeneous CAVs by assigning distinct configurations to each client for image resolution, batch size, dataset size, loaded from configuration files at runtime. This enables a sensitivity evaluation to hardware and data diversity, providing guidance for deployment in mixed environments. 

Our design reflects FL using a flower framework that can simulate real-world CAVs uses Flower to simulate real-world CAV environments by combining heterogeneous client capabilities, resource monitoring, and multi-model support. This produces reproducible and deployment relevant evaluation results for federated object detection models.

\section{\textbf{Experiment Setup}}
\label{sec:experiment_setup}

In this section, we describe the datasets, outline the FL client partitioning strategies, and present the evaluation metrics used to assess both accuracy and system efficiency. 

\subsection{\textbf{DATA}}

We used KITTI \cite{geiger2013vision}, BDD100K \cite{yu2020bdd100k} and nuScenes \cite{caesar2020nuscenes} datasets, which were selected for their diversity and relevance to real-world driving scenarios. These data sets capture a wide variety of conditions, including varying lighting, weather, and complex urban scenes. They also reflect practical challenges such as class imbalance and non-IID data distributions, making them well-suited for evaluating FL for object detection. A summary of their characteristics is provided in Table~\ref{tab:datasets}. To simulate non-IID data distributions, we partitioned KITTI and BDD100K datasets by unevenly distributing object classes across clients. In both datasets, certain clients contain higher proportions of frequent categories, while others are relatively sparse. For BDD100K, we excluded the train class as its limited presence did not contribute to model learning. 

\begin{table}[h!]
\centering
\renewcommand{\arraystretch}{1.2} 
\setlength{\tabcolsep}{3pt} 
\begin{tabular}{lccccc} 
\hline
\textbf{Dataset} & \textbf{Classes} & \textbf{Resolution} & \textbf{Weather} & \textbf{Day/Night} & \textbf{Location} \\ \hline
KITTI & 8 & 1242×375px & No & No & Urban \\ 
BDD100K & 10 & 1280×720px & Yes & Yes & Urban \\
nuScenes & 24 & 1600×1200px & Yes & Yes & Urban \\  \hline
\end{tabular}
\caption{Comparison of Datasets for Object Detection} 
\label{tab:datasets}
\vspace{-0.3cm}
\end{table}

The KITTI data set was split into four clients(C1, C2, C3, C4) as shown in Table~\ref{tab:Client_KITTI}) mainly due to its smaller volume. The BDD100K was divided into eight clients(C1, C2, C3, C4, C5, C6, C7, C8) to capture finer partitioning, resulting in greater heterogeneity between clients and enabling a more rigorous evaluation of cross-client variability and scalability, thereby better reflecting real-world CAV settings.

\begin{table}[!htbp]
\centering
\footnotesize   
\setlength\tabcolsep{4pt}
\begin{tabular}{|l|c|c|c|c|}
\hline
\textbf{Class} & \textbf{C1} & \textbf{C2} & \textbf{C3} & \textbf{C4} \\ \hline
Car     & 11508 & 5920 & 2925 & 2823 \\ \hline
Van    & 1173 & 615 & 264 & 280 \\ \hline
Truck    & 434 & 228 & 105 & 114 \\ \hline
Pedestrian     & 1814 & 934 & 425 & 426 \\ \hline
Person Sitting    & 117 & 29 & 7 & 17 \\ \hline
Cyclist     & 636 & 321 & 170 & 162 \\ \hline
Tram    & 210 & 92 & 28 & 92 \\ \hline
Misc  & 395 & 207 & 91 & 89 \\ \hline
\end{tabular}
\caption{Count of each class KITTI}
\label{tab:Client_KITTI}
\vspace{-0.3cm}
\end{table}

\begin{table}[!htbp]
\centering
\footnotesize  
\setlength\tabcolsep{2pt}
\begin{tabular}{|l|c|c|c|c|c|c|c|c|}
\hline
\textbf{Class} & \textbf{C1} & \textbf{C2} & \textbf{C3} & \textbf{C4} & \textbf{C5} & \textbf{C6} & \textbf{C7} & \textbf{C8} \\ \hline
Pedestrian      & 46279 & 22441 & 11381 & 5805 & 2690 & 1385 & 703 & 665 \\ \hline
Rider      & 2325 & 1088 & 577 & 291 & 138 & 48 & 20 & 30 \\ \hline
Car   & 356110 & 178724 & 88719 & 45365 & 21885 & 11284 & 5721 & 5403 \\ \hline
Truck       & 14909 & 7507 & 3705 & 1935 & 976 & 476 & 228 & 235 \\ \hline
Bus   & 5780 & 2968 & 1441 & 727 & 347 & 214 & 105 & 90 \\ \hline
Motor       & 1449 & 736 & 403 & 225 & 118 & 32 & 17 & 22 \\ \hline
Bike & 3820 & 1698 & 926 & 446 & 164 & 76 & 35 & 45 \\ \hline
Traffic Light  & 92792 & 46558 & 23342 & 11796 & 5659 & 3040 & 1495 & 1435 \\ \hline
Traffic Sign   & 120510 & 59410 & 30131 & 14896 & 7307 & 3812 & 1787 & 1833 \\ \hline
\end{tabular}
\caption{Count of each class BDD100K}
\label{tab:Client_BDD}
\vspace{-0.5cm}
\end{table}

The nuScenes dataset, which offers a larger number of categories and multimodal inputs, was divided into four subsets with a 50\%, 25\%, 12.5\%, and 12.5\% allocation across clients. This asymmetric split mirrors the availability of heterogeneous data across connected vehicles. Four rarely observed categories, pedestrian wheelchair, ego vehicle, emergency police, and emergency ambulance, were excluded because they did not provide sufficient samples for model training.

Although nuScenes also includes weather annotations, its coverage across scenarios and overall image volume are more limited compared to BDD100K. Therefore, we only conducted initial experiments on nuScenes, whereas the primary contributions focus on KITTI and BDD100K of our experiments, we provide detailed splits for these data sets in Tables~\ref{tab:Client_KITTI} and~\ref{tab:Client_BDD}.

To evaluate the performance of the model under diverse environmental conditions, we used the predefined weather and lighting annotations from the BDD100K dataset. Five weather categories were considered, clear, overcast, cloudy, rainy, and snowy. For lighting, we focused on the two dominant conditions, daytime and nighttime. As shown in Table~\ref{tab:lighting_weather}, each weather condition contains a substantial number of labeled images across lighting scenarios.

\begin{table}[htbp]
\centering
\renewcommand{\arraystretch}{1.2}
\begin{tabular}{lrrrrrr}
\toprule
\textbf{Condition} & \textbf{Clear} & \textbf{Overcast} & \textbf{Cloudy} & \textbf{Rainy} & \textbf{Snowy}  \\
\midrule
\textbf{Daytime}     & 14{,}218  & 8{,}590 & 4{,}900 & 2{,}930 & 3{,}284  \\
\textbf{Dawn/Dusk}   & 2{,}314   & 1{,}329 & 665     & 384     & 510      \\
\textbf{Night}       & 26{,}158  & 90     & 54      & 2{,}494 & 2{,}522  \\
\midrule
\textbf{Total}       & 42{,}690  & 10{,}009 & 5{,}619 & 5{,}808 & 6{,}316  \\
\bottomrule
\end{tabular}
\caption{Image distribution across lighting and weather}
\label{tab:lighting_weather}
\vspace{-0.5cm}
\end{table}

For each weather and lighting condition, the images were randomly partitioned into five subsets of clients with distributions of client 1 (30\%), client 2 (25\%), client 3 (20\%), client 4 (15\%) and client 5 (10\%). Specifically, the subset assigned to Client 3 (20\%) was used consistently as the test set, ensuring a uniform evaluation in all environmental scenarios.

\vspace{-0.3cm}
\subsection{\textbf{Evaluation Metrics}}

We evaluated object detection models using both accuracy and system efficiency measures listed below.

\subsubsection{\textbf{Mean Average Precision (mAP)}} It evaluates model’s ability to correctly localize and classify objects by integrating precision recall curves across different Intersection over Union (IoU) thresholds. We report mAP at IoU=0.5, higher mAP values indicate more accurate detection performance.

\subsubsection{\textbf{GPU Memory Consumption (GiB)}} This metric captures the peak memory footprint during training and inference. Since modern object detectors often operate on resource-constrained CAVs, memory profiling provides information on the scalability and feasibility of the model for deployment.

\subsubsection{\textbf{GPU Utilization (\%)}} GPU utilization measures the proportion of available GPU resources that are actively used during execution. It reflects how efficiently the detector uses hardware, which is essential for evaluating throughput and ensuring a fair comparison between models.

\subsubsection{\textbf{Power Consumption (W)}} Power usage is monitored in watts(W) to quantify the real-time energy demand of the GPU. In energy-sensitive domains such as CAVs, low power consumption without compromising accuracy is critical for sustainable deployment.

\subsubsection{\textbf{Inference Latency (ms)}} Latency represents the average time required to process a single input image. Measured in milliseconds(ms), it directly determines the suitability of the detector for real-time CAVs, where low-latency responses are mandatory for safety-critical decision making.

All metrics were collected within our FL framework. mAP was recorded during evaluation, while GPU memory, utilization, and power consumption were monitored using \textit{pynvml}, \textit{psutil}, and \textit{nvidia-smi}. Inference latency was measured as the average forward-pass time per image. This setup ensured a consistent assessment of both accuracy and system efficiency.



\section{\textbf{Results}}
\label{sec:results}

\begin{table*}[!htbp]
\small
    \centering
    \renewcommand{\arraystretch}{1.2}
    \begin{tabular}{|l|ccc|ccc|ccc|ccc|}
        \hline
        \multirow{2}{*}{\diagbox[width=6em,height=2.7em,dir=SW]%
        {\hspace{-0.4em}\textbf{Dataset}}%
        {\textbf{Model}}}
        & \multicolumn{3}{c|}{\textbf{Centralized}} 
        & \multicolumn{3}{c|}{\textbf{FedAvg}} 
        & \multicolumn{3}{c|}{\textbf{FedProx}} 
        & \multicolumn{3}{c|}{\textbf{FedAsync}}\\
        \cline{2-13}
        & v5 & v8 & v11 
        & v5 & v8 & v11 
        & v5 & v8 & v11
        & v5 & v8 & v11\\
        \hline
        KITTI     
        & 78.4 & 81.6 & 77.6 
        & 82.9 & 87.5 & 84.0 
        & 83.4 & \textbf{87.9} & 84.1 
        & 78.3 & 82.0 & 80.1    \\
        BDD100K  
        & 58.9 & 61.4 & 60.2 
        & 59.3 & \textbf{61.5} & 60.5 
        & 59.6 & \textbf{61.5} & 60.7 
        & 42.3 & 45.7 & 44.1     \\
        nuScenes  
        & 53.2 & 56.3 & 55.9  
        & 57.3 & 60.6 & 59.4   
        & 57.7 & \textbf{60.8} & 59.9
        & 38.6 & 41.8 & 41.7 \\
        \hline
    \end{tabular}
    \caption{mAP comparison of centralized and federated learning across YOLO versions.}
    \label{tab:yolomap}
    \vspace{-0.3cm}
\end{table*}

To focus on deployment scenarios, the experiments were conducted in stages, each designed to isolate the most informative conditions. First, KITTI, BDD100K, and nuScenes data sets were evaluated in centralized and FL to determine which best represented structured environments and diverse conditions; KITTI and BDD100K were chosen for the remaining analysis. After this selection, image resolutions and batch sizes were varied to assess trade-offs between accuracy and resource usage, followed by measurements of GPU memory, computation usage, power consumption, inference time, training time, and scalability. These evaluations on YOLO models were followed by tests on client dropout and variations in image resolution across clients. Finally, the Deformable DETR was evaluated in centralized and federated settings for day and night scenarios only, as earlier results indicated that the strongest performer had already been identified for detailed weather conditions experiments. This sequence ensured that each stage was purposeful and built on the previous one.

To evaluate federated object detection, centralized training was conducted for 10 epochs, while FL experiments used 10 communication rounds, with 3 local epochs per round. FedAvg and FedProx were implemented in a synchronous manner, where all clients participate in each round, whereas FedAsync used 10 rounds but followed an asynchronous protocol. 

\vspace{-0.35cm}
\subsection{\textbf{Centralized and Federated Detection Performance}}

Table~\ref{tab:yolomap} presents the mean Average Precision (mAP) of YOLOv5, YOLOv8, and YOLOv11 across centralized and FL paradigms on the KITTI, BDD100K, and nuScenes datasets.

Centralized learning serves as the baseline. On KITTI dataset, YOLOv5, YOLOv8, and YOLOv11, achieved mAP of 78.4\%, 81.6\%, and 77.6\% respectively. On BDD100K, their scores were 58.9\%, 61.4\%, and 60.2\%, while on nuScenes they reached 53.2\%, 56.3\%, and 55.9\%. Object detection accuracy declines as datasets progress from structured KITTI scenes to more diverse BDD100K and rare events in nuScenes, reflecting the increasing complexity of real-world CAVs. 

For FL, FedAvg (standard synchronous method) with YOLOv5, YOLOv8, and YOLOv11 achieved mAP scores of 82.9\%, 87.5\%, and 84.0\% on KITTI, 59.3\%, 61.5\%, and 60.5\% on BDD100K; and 57.3\%, 60.6\%, and 59.4\% on nuScenes, respectively. FedProx introduces a proximal term to reduce the impact of client drift, marginal gains over FedAvg across all datasets: on KITTI, scores reach 83.4\%, 87.9\%, and 84.1\%, on BDD100K they are 59.6\%, 61.5\%, and 60.7\%, and on nuScenes 57.7\%, 60.8\%, and 59.9\%. By introducing minimal proximal term, FedProx achieved slight improvements over FedAvg, stabilizing updates under non-IID conditions. Our results show that synchronous FL can achieve accuracy comparable to or higher than centralized training. 


FedAsync achieves 78.3\%, 82.0\%, and 80.1\% mAP on KITTI, 42.3\%, 45.7\%, and 44.1\% on BDD100K, and 38.6\%, 41.8\%, and 41.7\% on nuScenes for YOLOv5, YOLOv8, and YOLOv11, respectively. Asynchronous learning progresses faster by incorporating updates from faster clients. This accelerates training, meaning contributions are dominated by faster clients, and the aggregated model tends to over represent their data while under utilizing contributions from slower clients, noticeable in complex datasets like BDD100K and nuScenes


Our results show that synchronous learning with FedAvg and FedProx maintains strong performance. While asynchronous learning converges faster with fewer updates, it requires refined strategies to balance client contributions. To ensure a fair comparison, this study was limited to ten updates and revealed that model complexity matters. 
Based on our findings, we adopt the synchronous approach in subsequent analysis to capture maximum detection performance while evaluating system-level metrics, client participation dynamics, and environmental conditions. 
We focus on KITTI and BDD100K, together reflect structured urban driving and diverse weather and lighting scenarios essential for evaluation.

\vspace{-0.35cm}
\subsection{\textbf{Client Resource Profiling in FL}}
In CAVs, vehicles have heterogeneous hardware, with differences in computing power, memory, and energy constraints creating system-level challenges for the deployment of FL. 

\subsubsection{\textbf{Parameter Evaluation}}

We evaluated the impact of image resolution and batch size on performance and resource utilization. Resolutions of 320, 640, and 960 were tested with a batch size of 32 to capture scaling effects, while additional runs at 960 with batch sizes of 16, 8, and 4 assessed the impact of smaller batches under high-resolution inputs. These configurations were designed to approximate hardware limitations in heterogeneous vehicles, where memory and compute capacity vary significantly.


\begin{table}[ht]
\scriptsize
\setlength{\tabcolsep}{2pt}
\renewcommand{\arraystretch}{1.0}
\centering
\begin{tabular}{cccccccccc}
\toprule
\textbf{Parameter} & \textbf{Value} & \textbf{Dataset} 
& \multicolumn{2}{c}{\textbf{v5x6u}} 
& \multicolumn{2}{c}{\textbf{v8x}} 
& \multicolumn{2}{c}{\textbf{v11x}} \\
\cmidrule(lr){4-5} \cmidrule(lr){6-7} \cmidrule(lr){8-9}
& & & FedAvg & FedProx & FedAvg & FedProx & FedAvg & FedProx \\
\midrule
\multirow{6}{*}{Image Size} 
& \multirow{2}{*}{320} & KITTI & 69.7 & 70.1 & 71.2 & 71.4 & 67.8 & 68.2 \\
&                     & BDD   & 40.3 & 40.5 & 41.2 & 41.6 & 40.9 & 41.1 \\
\cmidrule{2-9}
& \multirow{2}{*}{640} & KITTI & 82.9 & 83.4 & 87.5 & 87.9 & 84.0 & 84.1 \\
&                     & BDD   & 58.9 & 59.6 & 61.5 & 61.5 & 60.5 & 60.7 \\
\cmidrule{2-9}
& \multirow{2}{*}{960} & KITTI & 87.0 & 87.2 & 90.2 & 90.6 & 87.5 & 87.7 \\
&                     & BDD   & 61.9 & 62.2 & 66.7 & 66.9 & 64.8 & 65.1 \\
\midrule
\multirow{6}{*}{Batch Size} 
& \multirow{2}{*}{16} & KITTI & 86.9 & 87.2 & 90.5 & 90.7 & 88.3 & 88.5 \\
&                     & BDD   & 61.1 & 61.2 & 65.7 & 65.9 & 63.9 & 64.1 \\
\cmidrule{2-9}
& \multirow{2}{*}{8}  & KITTI & 86.8 & 87.1 & 89.5 & 89.7 & 86.8 & 87.0 \\
&                     & BDD   & 59.2 & 59.6 & 64.1 & 64.2 & 61.8 & 61.9 \\
\cmidrule{2-9}
& \multirow{2}{*}{4}  & KITTI & 86.0 & 86.3 & 88.9 & 89.1 & 86.2 & 86.3 \\
&                     & BDD   & 57.2 & 57.4 & 62.9 & 63.6 & 59.3 & 59.5 \\
\bottomrule
\end{tabular}
\caption{Mean Average Precision (mAP) of YOLO variants under FedAvg and FedProx on KITTI and BDD100K with varying image resolutions and training batch sizes.}
\label{tab:Image_Batch}
\vspace{-0.3cm}
\end{table}

Table~\ref{tab:Image_Batch} presents the effect of image resolution and batch size on detection performance under FedAvg and FedProx. For image resolution, smaller input sizes lead to reduced accuracy, with KITTI achieving 69.7–71.4\% and BDD100K 40.3–41.6\% at 320×320. Increasing the resolution to 640×640 improves the results to 82.9–87.9\% on KITTI and 58.9–61.5\% on BDD100K, while 960×960 achieves the highest mAP of 87–90.6\% and 61.9–66.7\% on the FedAvg, FedProx, respectively. This progression reflects that higher-resolution inputs capture finer details, improving detection accuracy. 

For batch size evaluation, the YOLO models maintain consistently high detection accuracy on KITTI across all configurations, showing 86.9-90.7\% mAP in batch size 16, 86.8-89.7\% in batch size 8 and 86.0-88.9\% in batch size 4. However, BDD100K shows a clear performance decline as the batch size decreases. At batch size 16, mAP ranges from 61.1–65.7\%, while reducing to batch size 8 yields 59.2–64.2\%, and the smallest batch size of 4 drops to 57.2–63.6\%.

These results show that smaller batch sizes can produce noisier local updates, while higher resolutions consistently improve detection accuracy; in contrast, small batches or low resolutions degrade performance. The parameter selection shapes heterogeneous FL, with profiling showing that KITTI peaks at 960×960 with a batch size of 16, offering a reference to adapt training across vehicles. Adjusting these parameters provides deployment guidance to maintain accuracy under diverse memory and processing constraints in real-world CAVs.

\subsubsection{\textbf{Model Architecture}}

Table~\ref{tab:model-comparison} compares YOLOv5, YOLOv8, and YOLOv11 in terms of layers, parameters, GFLOPs, and trainable gradients. YOLOv5 has the largest architecture (155.45M parameters), YOLOv11 the smallest (56.88M) and YOLOv8 lies between (68.16M). These metrics provide a quick estimate of model size, computational cost, and potential communication overhead in FL.

\begin{table}[!htbp]
\centering
\begin{tabular}{|l|c|c|c|c|}
\hline
\textbf{Model} & \textbf{Layers} & \textbf{Parameters (M)} & \textbf{GFLOPs} & \textbf{Gradients (M)} \\
\hline
YOLOv5x6u  & 638 & 155.45 & 251.3  & 155.45 \\
YOLOv8x    & 365 & 68.16  & 258.2  & 68.16 \\
YOLOv11x   & 357 & 56.88  & 195.5  & 56.88 \\
\hline
\end{tabular}
\caption{Comparison of Model Architectures}
\label{tab:model-comparison}
\vspace{-0.3cm}
\end{table}

However, architectural metrics alone do not fully determine system-level resource requirements for deployment. Peak resource demand is influenced by activation sizes, feature map widths, and fused operations during training, which affect resource utilization. In heterogeneous CAVs, where edge devices differ widely in capacity, selecting models based solely on parameters or GFLOPs can lead to inefficient or infeasible deployments. Reliable model selection for FL must combine architectural analysis with empirical profiling to capture true system-level costs on diverse hardware.

\subsubsection{\textbf{GPU Memory}}

\begin{figure*}[t]
    \centering
    \includegraphics[width=1.0\textwidth]{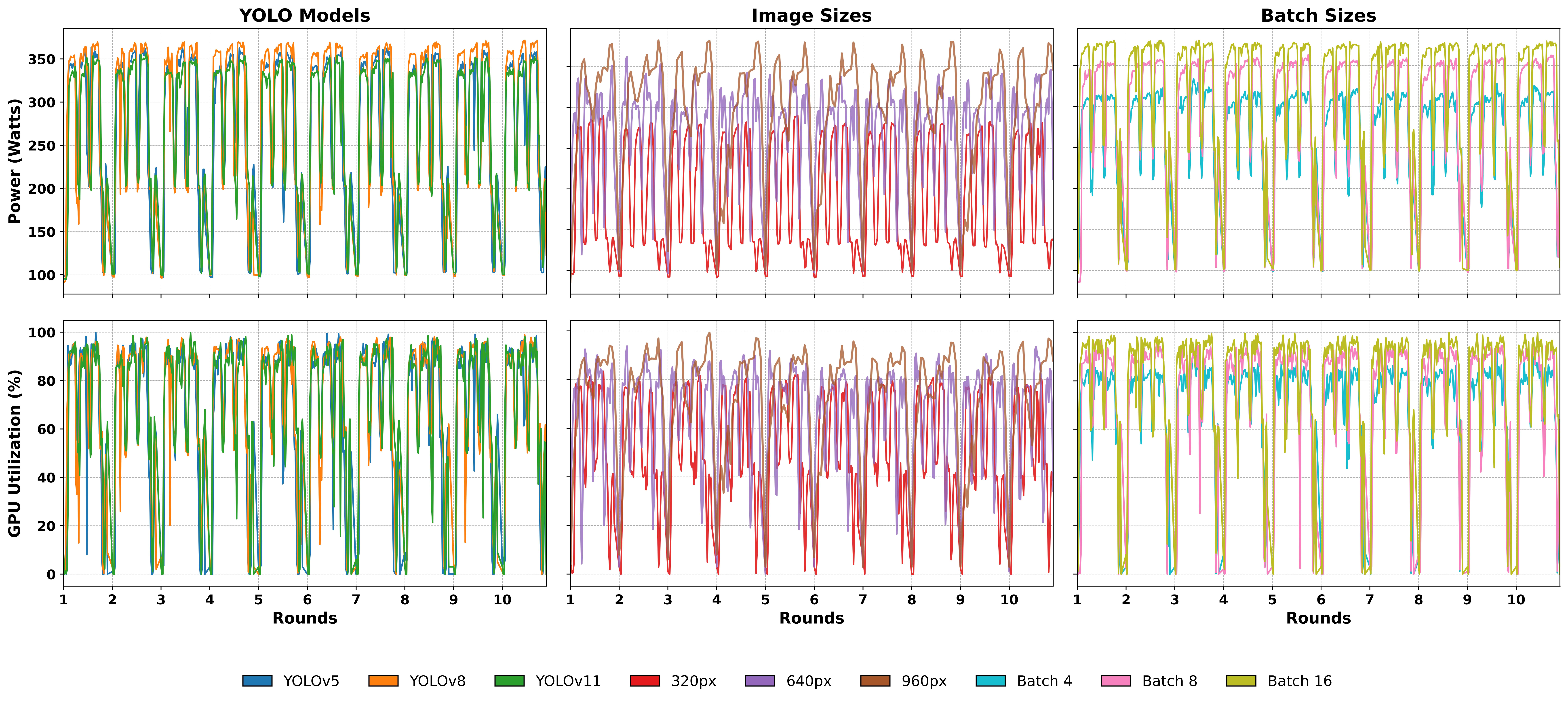}
    \caption{GPU and power profiles across YOLO models, image resolutions, and batch sizes}
    \label{fig:GPUPowerUtilisation}
    \vspace{-0.4cm}
\end{figure*}

GPU memory is a critical bottleneck in CAVs, where modules such as object detection, segmentation, tracking, and sensor fusion operate concurrently on hardware. Models exceeding available memory can delay or stop essential tasks, potentially compromising real-time perception and overall driving safety. Memory-aware model selection is therefore essential for reliable deployment in FL settings. Since training requires more memory, utilization, and power than inference, any GPU sufficient for training can also handle inference, therefore our analysis focuses on training.

\begin{figure}[H]
    \centering
    \includegraphics[width=0.4\textwidth]{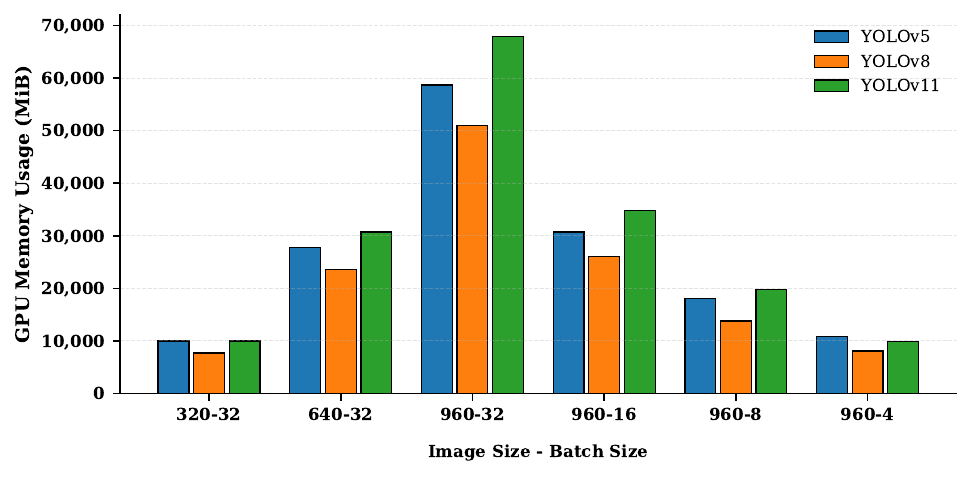}
    \caption{GPU memory usage of YOLO models}
    \label{fig:GPUImage}
    \vspace{-0.5cm}
\end{figure}

Figures~\ref{fig:GPUImage} presents the GPU memory usage of YOLOv5x6u, YOLOv8x, and YOLOv11x under synchronized FL, evaluated across three image resolutions (320, 640, and 960) at a batch size of 32. At 960×960, memory usage is highest, with 67379, 58675, and 50892 MiB and at 640×640, this drops to 30617, 27750, and 23552 MiB, and at 320×320 is 10340, 9113, and 7680 MiB, for YOLOv11, YOLOv5 and YOLOv8, respectively. The difference between models is clearly observed at 960×960, reflecting the impact of architectural complexity and tensor scaling. This profiling highlights that model and resolution specific memory evaluations are essential for deployment, as memory demand indicate which models can safely operate on resource-constrained vehicles.

A second experiment on batch size effects at 960×960. YOLOv11 requires 10035, 18841, and 35635.2 MiB, YOLOv5 uses 11090, 17920, and 31436 MiB, and YOLOv8 remains the most memory efficient at 8294, 16290, and 26726 MiB, at batch sizes 4, 8 and 16, respectively. Memory demand scales with batch size and model highlight the need for profiling to prevent OOM errors and ensure safe CAV deployment.

\subsubsection{\textbf{GPU Computation Utilization}}

GPU computation utilization is the active workload on the GPU during training in FL. Monitoring utilization helps identify peak computational demand and guide scheduling to prevent hardware overload. 

The GPU utilization patterns show a cyclical trend across 10 FL rounds, peaking during local training and dropping sharply during aggregation. The first subplot in Figure~\ref{fig:GPUPowerUtilisation} shows GPU utilization reaching 85–95\% during training, with YOLOv5, YOLOv8 and YOLOv11 showing similar trends. After each round, utilization drops to 0–10\%, indicating that GPUs remain largely idle during global aggregation and communication. The second sub-plot illustrates the effect of image resolution on GPU utilization. High-resolution inputs (960×960) have the highest peaks, reaching approximately 90–95\%, with gradual declines between epochs due to the extended processing of the larger tensors. Medium resolution inputs (640×640) produce peaks around 80–90\%, while low-resolution inputs (320×320) show utilization near 70–80\%. This trend confirms that lower image resolutions demand less computation, resulting in reduced GPU utilization.

The third subplot shows the impact of batch size. The larger batches (batch-16) achieve the highest utilization, around 90–95\%, followed by batch-8 with peaks near 85–90\%, and batch-4 fluctuating between 80–85\% with visible idle gaps. Larger batches increase utilization because more images are processed simultaneously, keeping the GPU fully occupied during forward and backward passes. Smaller batches, in contrast, complete processing faster, creating proportionally longer idle periods and leaving the GPU underutilized.

This analysis highlights utilization behavior and providing deployment guidance to align training workloads with vehicle GPU capacity. Understanding these patterns is essential for optimizing client participation and ensuring FL training coexists efficiently with other on-board processes in CAVs.


\subsubsection{\textbf{Power Consumption}}

\begin{figure*}[t]
    \centering
    \includegraphics[width=1.0\textwidth]{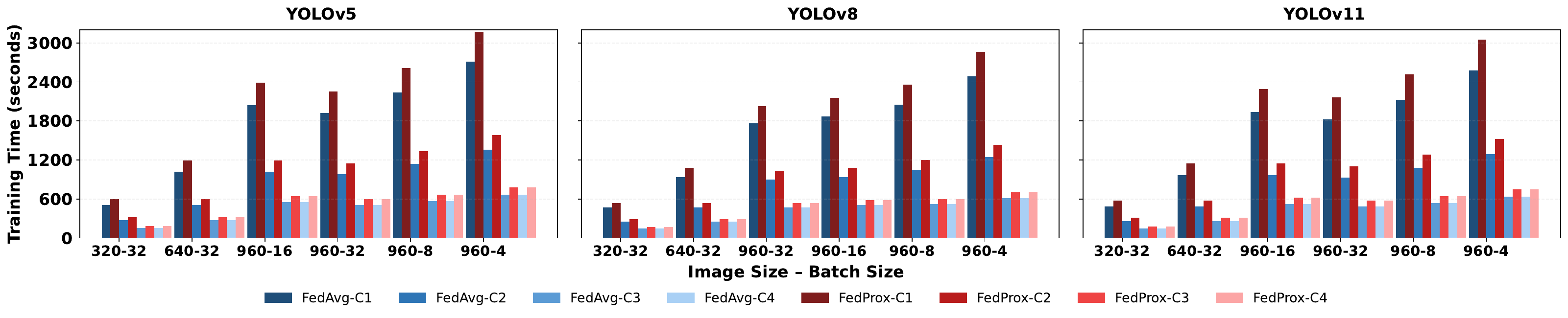}
    \caption{Training time comparison for YOLOv5, YOLOv8, and YOLOv11 in FL}
    \label{fig:Time}
    \vspace{-0.2cm}
\end{figure*}

Power consumption is critical for CAVs, as sustained high GPU draw accelerates battery depletion and limits operating time. Intensive training or inference workloads can also increase thermal output, potentially causing performance degradation or system shutdowns if not managed. Therefore, monitoring and analyzing power usage in FL provides insight into how training configurations affect energy demand, enabling resource-aware deployment that balances model performance with operational longevity.

The power consumption shows a cyclical pattern, peaking during local training and dropping during aggregation, consistent with the GPU utilization trends. Among the models, YOLOv11 consumes the least power, ranging from 325–350W, YOLOv5 is moderate at approximately 335–360W, and YOLOv8 reaches the highest levels at 350W-375W. For image resolution, 320×320 inputs consume 250–280W, 640×640 inputs rise to 300–350W, and 960×960 inputs exceed 350W, confirming that higher resolutions require more power. The batch size shows clearer impact: batch-4 remains around 300–320W, batch-8 increases to 340–350W, and batch-16 peaks near 350–375W, indicating that larger batches fully utilize the GPU and draw the most power. Such analysis is essential for CAVs to guide energy-aware model selection and maintain operational time and safety.


\subsubsection{\textbf{Training Time}}
Training time is critical in FL for CAVs because vehicles cannot operate continuously and must complete model updates within a limited operating time. As shown in Figure~\ref{fig:Time}, for C1 with the 640×32 configuration YOLOv5, YOLOv8, and YOLOv11 require 1,019.58s, 936.00s, and 969.42s per 10 rounds under FedAvg, respectively, while FedProx introduces slightly higher times of 1,194.05s, 1,076.40s, and 1,147.12s. Similar trends are observed for other clients, where FedProx consistently incurs a small overhead due to the proximal term. Across all configurations, higher image resolutions substantially increase training time, for example, YOLOv8 image resolution strongly impacts training time: 320×32 completes in 468s, 640×32 rises to 936s, and 960×32 peaks at 1,764s. While smaller batch sizes extend training, where 960×960 with batch-4 requires 2,488s, batch-8 reduces to 2,052s, batch-16 to 1,872.00s, and batch-32 achieves the fastest 1,764.00s.


These results show that model architecture and aggregation strategy directly influence training duration, making careful selection essential to meet operational constraints. Training-time analysis is critical for CAV deployments as it guides scheduling federated updates without compromising vehicle availability or real-time perception.

\subsubsection{\textbf{Inference}}

\begin{figure*}[t]
    \centering
    \includegraphics[width=\textwidth, height=0.13\textheight]{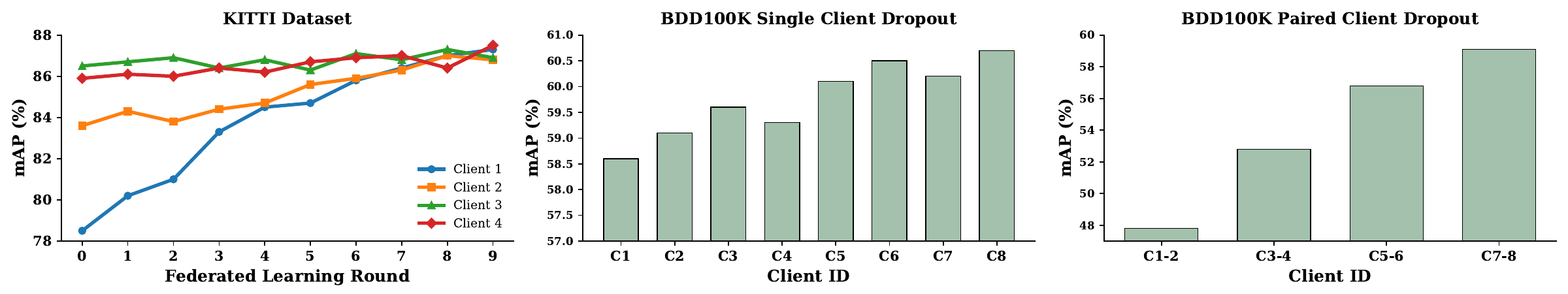}
    \caption{Effect of client dropout on mAP for KITTI and BDD100K}
    \label{fig:ClientDropout}
    \vspace{-0.3cm}
\end{figure*}

Inference speed is critical for CAVs, as perception and decision making in real time directly affect safety and response. 

\begin{table}[htbp]
\vspace{-0.25cm}
\renewcommand{\arraystretch}{1.0}
\centering
\label{tab:inference_latency}
\begin{tabular}{llccc}
\toprule
\textbf{Dataset} & \textbf{Image Size} & \textbf{YOLOv5} & \textbf{YOLOv8} & \textbf{YOLOv11} \\
\midrule
\multirow{3}{*}{KITTI} 
 & 320 & 0.4 & 0.5 & 0.6 \\
 & 640 & 0.9 & 1.1 & 1.2 \\
 & 960 & 1.7 & 1.9 & 2.1 \\
\midrule
\multirow{3}{*}{BDD100K} 
 & 320 & 0.7 & 0.8 & 1.0 \\
 & 640 & 1.3 & 1.6 & 1.7 \\
 & 960 & 2.4 & 2.6 & 3.1 \\
\bottomrule
\end{tabular}
\caption{Inference(ms) across image resolutions}
\vspace{-0.25cm}
\end{table}


All images were resized to fixed input resolutions (320×320, 640×640, and 960×960 at batch size 32), the inference time still increased with larger input sizes because higher‑resolution images contain more pixels, requiring more computations per frame. In addition, the BDD100K dataset consistently exhibited longer inference times than KITTI, even at the same resized resolution, due to its larger original image dimensions and denser scenes with more objects, which result in more predicted bounding boxes to process.

\subsubsection{\textbf{Testing Scalability}}
To assess how the quantity of data per client affects object detection performance in a FL setting, we conducted experiments using 60 clients with the KITTI dataset. In the first configuration, the data set was evenly partitioned between clients and each client received a disjoint subset for local training. This configuration resulted in an mAP@0.5 of 69.7\%, highlighting that limited local data per client restricts learning capacity and global convergence. 

To address this, we designed an overlapping data partition scheme. Each client received five consecutive partitions of the data set (e.g. C1: partitions 1–5, C2: 2–6, ..., C60: 56–60), thus increasing the local data volume five times (with some repetition). Despite redundancy, this expanded set-up improved mAP@0.5 to 88.1, demonstrating that increased local data volume, independent of the number of clients, significantly enhances detection accuracy. This finding emphasizes the need to balance the number of clients with the richness of data per client in FL scenarios for CAVs.

To evaluate FL scalability, we tested 60 and 800 clients, both were feasible using the Flower framework. As Beutel et al. \cite{beutel2020flower} demonstrated that Flower can scale to client pools of up to 15 million and support up to 1,000 concurrent clients per round, observing slower convergence beyond that point.

\vspace{-0.25cm}
\subsection{\textbf{Global Model Sensitivity to Client Variability}}
Building on the client-level trade-offs associated with irregular participation, the following analysis shifts to the global model-level perspective. 

\subsubsection{\textbf{Client Dropout}}

In CAVs, intermittent connectivity from limited coverage, mobility, or disconnections prevents full participation in every training round, slowing convergence, and reducing accuracy in FL. Evaluating performance under these conditions is essential to ensure robustness in deployment.


We configured the KITTI dataset to simulate client additions and dropouts in each round. As shown in Figure~\ref{fig:ClientDropout}, KITTI results, round-wise performance, show that global mAP increases steadily as rounds progress, but individual round contributions vary in importance. C1 exhibits mAP values of 78.5\%, 80.2\%, 81.0\%, 83.3\%, 84.5\%, 84.7\%, 85.8\%, 86.4\%, 87.0\%, and 87.3\% across rounds 1–10, while clients 2, 3, 4 achieve similar performance, ending at 86.8\%, 86.9\%, and 87.5\%. Each round incrementally improves the global model, 
but missing client updates can slow convergence by reducing data diversity. 
The steadily increasing mAP across rounds illustrates the convergence behavior even under client absence. 


For the BDD100K dataset, the impact of dropout is more evident when comparing single and dual-client absence. Single-client dropouts result in mAP values of 58.6\% (C1), 59.1\% (C2), 59.6\% (C3), 59.3\% (C4), 60.1\% (C5), 60.5\% (C6), 60.2\% (C7), and 60.7\% (C8), indicating only marginal performance reductions, with clients C1 and C2 having a slightly more noticeable effect due to their larger data volumes. Given that single-client absence resulted in only marginal differences, dual-client scenarios were further explored to capture more pronounced impacts on convergence and accuracy. The dual-client dropouts lead to a significant decline in accuracy: 47.8\% (C1-2), 52.8\% (C3-4), 56.8\% (C5-6), and 59.1\% (C7-8). The greatest degradation occurs when C1-2 are absent, as these clients contribute the majority of the data, while the impact diminishes for C7-8, which hold smaller partitions.

Our results show that the effect of client dropout depends on the dataset scale. In KITTI, losing any client causes noticeable accuracy fluctuations since each partition provides a significant contribution to learning. BDD100K is more distributed, so single client dropouts lead to only marginal accuracy changes, though multi-client dropouts cause substantial degradation. This underscores the importance of broad client participation for maintaining global accuracy and highlights the challenge of large-scale client unavailability in real deployments. 

\subsubsection{\textbf{Client Heterogeneity}}

\begin{figure*}[t]
    \centering
    \includegraphics[width=\textwidth, height=0.175\textheight]{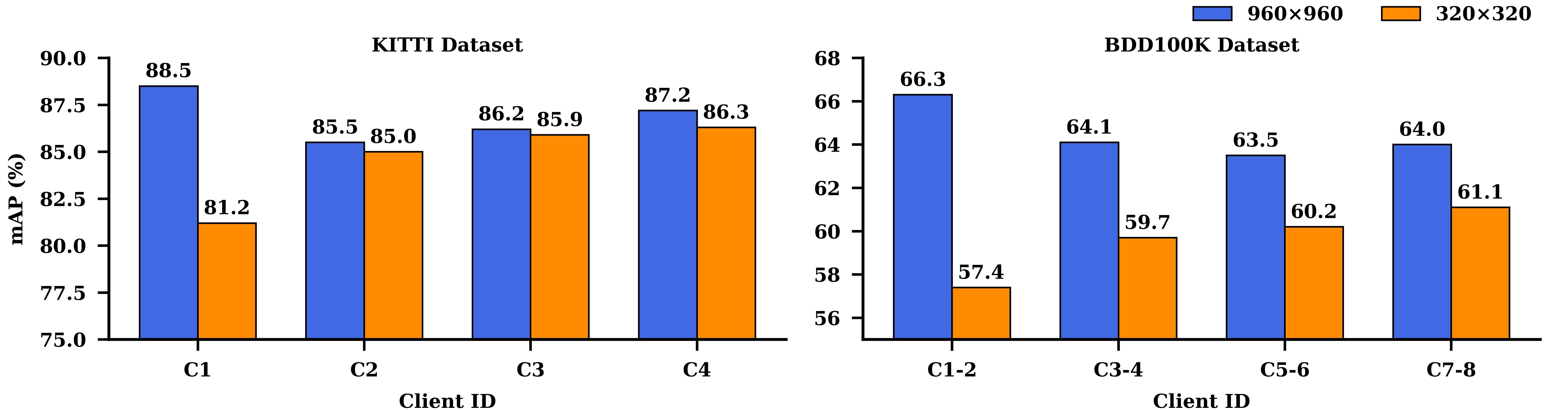}
    \caption{Client resolution heterogeneity and its effect on detection performance}
    \label{fig:ClientHetero}
    \vspace{-0.3cm}
\end{figure*}

CAVs differ in onboard sensor resolution, creating inherent heterogeneity in perception. While high-resolution sensing is possible, limited computational resources often require reduced resolutions, adding further imbalance. To simulate this, we design an FL experiment with non-uniform image resolutions and non-i.i.d. data. In this setup, the blue bars indicate a client operating at 960×960 while others use 640×640, and the orange bars indicate a client at 320×320 while others remain at 640×640, with all clients using a batch size of 32.


For the KITTI dataset, C1 at 960×960 achieves 88.5\% mAP, compared to 81.2\% when set to 320×320. C2 reaches 85.5\% vs. 85.0\%, C3 achieves 86.2\% vs. 85.9\%, and C4 shows 87.2\% vs. 86.3\%. This explicitly shows that higher-resolution configurations provide the largest gain for C1, which holds the most data, while the effect diminishes for smaller clients. From the client-dropout analysis, a substantial change in global accuracy was observed when two clients were removed. Motivated by this, the heterogeneity evaluation was also configured with two clients at high resolution (960×960) and the remaining clients at 640×640. In BDD100K, C1‑2 achieved 66.3\% at 960×960 and 57.4\% at 320×320, C3‑4 reached 64.1\% and 59.7\%, C5‑6 achieved 63.5\% and 60.2\%, and C7‑8 obtained 64.0\% and 61.1\%, at high and low resolutions, respectively.


These experiments demonstrate that heterogeneous client configurations affect global model accuracy, with high capacity clients offering the greatest benefit. This analysis provides practical guidance for real‑world deployment, where vehicle resources vary and careful configuration such as prioritizing high-resolution or high-data-volume clients during training, while still including low-capacity clients to maintain fairness and diversity. Such balance can improve model reliability under unpredictable conditions.

\vspace{-0.2cm}

\subsection{\textbf{Transformer Based Object Detection Model}}
Following the YOLO-based evaluation, we included Deformable DETR as a transformer-based detector, given its popularity and broad adoption compared to other less widely used transformer variants. We evaluated Deformable DETR on the KITTI and BDD100K datasets in centralized and FL settings, focusing on trade-offs relevant to CAV deployment. Centralized training used 50 epochs, while FL consisted of 30 communication rounds with 3 local epochs per round.

In CAV systems, variability in camera resolutions across vehicles complicates consistent object detection. Detection accuracy is highly sensitive to viewing perspectives: Zeng et al.\cite{zeng2024ars} show that square-shaped objects are more robust to angular deviations, while elongated ones degrade significantly. Preserving aspect ratios during image preprocessing is therefore critical to avoid distortion and to maintain reliable feature extraction across heterogeneous driving scenes.

The deformable DETR applies an adaptive resizing strategy that maintains geometric consistency by scaling the shorter edge of each image to 800px and capping the longer edge at 1333px. This approach preserves the original aspect ratio, which is essential for reliable perception in varied driving environments. The native resolutions of KITTI (1242×375px) and BDD100K (1280×720px) differ significantly, after resizing, BDD100K images retain more vertical detail and a higher overall pixel count compared to KITTI, since their short edge is already close to 800 px and undergoes less downscaling. As a result, BDD100K images require more GPU memory during training, which limits feasible batch sizes. Accordingly, we used a batch size of 32 for KITTI and 16 for BDD100K to fit within the available GPU memory budget.

\begin{table}[h!]
\centering
\begin{tabular}{lcccc}
\toprule
\textbf{Dataset} & \textbf{Batch Size} & \textbf{GPU Memory} & \textbf{Centralized} & \textbf{FedAvg} \\
\midrule
KITTI  & 32 & 95,130 MiB  & 68.07\% & 74.67\% \\
BDD100K  & 16 & 87,320 MiB  & 53.56\% & 58.04\% \\
\bottomrule
\end{tabular}
\caption{Deformable DETR detection performance}
\label{tab:deformable_results}
\vspace{-0.5cm}
\end{table}

As shown in Table ~\ref{tab:deformable_results} the centralized training setting, the KITTI data set achieved a mAP of 68.07\%. In particular, federated training outperformed this baseline with an mAP of 74.67\%. For this setup, KITTI (batch size 32) consumed approximately 95,130 MiB of GPU memory. On the BDD100K dataset, centralized training achieved an mAP of 53.56\%, while FL achieved an mAP of 58.04\%. Training BDD100K (batch size 16) required around 87,320 MiB of GPU memory. Due to the prohibitively high computational cost of the Deformable DETR model. Training on the BDD100K dataset in FL setup, which required more than 72 hours to complete the training, therefore, we discontinued further experiments with Deformable DETR after the illumination study, reflecting its limited suitability for resource-constrained CAV deployment.

The aspect ratio-preserving resizing strategy employed by Deformable DETR facilitates more realistic and robust object detection across heterogeneous visual inputs. However, this comes with non-uniform memory demands and variable batch size constraints across datasets. These tradeoffs must be carefully managed to balance detection accuracy, efficiency, and scalability in resource-constrained, safety-critical CAVs.

\subsection{\textbf{Environmental Impact on Detection Performance}}

\subsubsection{\textbf{Light Conditions}}

A key challenge in federated object detection for CAVs arises from lighting variations between day and night, which create non-IID data as illumination alters image characteristics. 
To assess the impact of lighting-induced domain shifts, we trained clients separately on day-only and night-only datasets and evaluated model performance under both in-domain and cross-domain test conditions. 


\begin{table}[ht]
\scriptsize
\setlength{\tabcolsep}{3pt}
\centering
\small
\begin{tabular}{lcccccccc
}
\toprule
\textbf{Train} 
& \multicolumn{4}{c}{\textbf{Day Clients}} 
& \multicolumn{4}{c}{\textbf{Night Clients}} 
\\
\cmidrule(lr){2-5} \cmidrule(lr){6-9} 
\textbf{Test}  
& v5 & v8 & v11 & Def-DETR 
& v5 & v8 & v11 & Def-DETR 
\\
\midrule
Day   
& 57.2 & 62.5 & 61.7 & 63.85 
& 46.4 & 48.6 & 47.1 & 41.94 
\\
Night 
& 48.8 & 51.3 & 49.8 & 42.35 
& 50.5 & 54.2 & 53.1 & 49.77 
\\
\bottomrule
\end{tabular}
\caption{Performance (mAP \%) of YOLO (v5, v8, v11) and Deformable DETR under different client training conditions across day and night test scenarios.}
\label{tab:lighting}
\vspace{-0.5cm}
\end{table}

As shown in Table~\ref{tab:lighting}, all models achieved the highest performance when training and testing conditions were matched. When trained and tested under the same lighting conditions (day-to-day or night-to-night), models achieve their highest performance. The deformable DETR shows the strongest in domain performance with 63.85\% mAP for day-to-day and 49.77\% for night-to-night. YOLOv8 also shows robust performance with 62.5\% (day-to-day) and 52.9\% (night-to-night). However, in cross-domain settings, performance decreases. For example, the deformable DETR trained on day and tested on night drops sharply to 42.95\%, indicating high sensitivity to lighting changes. In contrast, YOLOv11 achieved 59.2\% (day to day) to 49.8\% (day to night) and from 50.5\% to 47.1\% (night to day). YOLOv5 and YOLOv8 follow similar patterns, but YOLOv5 under night-to-day shows the lowest score among all settings (46.4\%). These results highlight that object detectors trained without accounting for lighting variability exhibit reduced generalization due to domain mismatch between clients, limiting their effectiveness in real-world deployments where such shifts are common.

\subsubsection{\textbf{Weather Conditions}}

Environmental factors such as weather significantly affect the performance of object detection models. Rain, snow, and overcast conditions reduce visibility and lower detection accuracy, making weather aware evaluation essential. Using BDD100K pre-annotations, we assess two aspects: in-domain performance (trained and tested in the same weather) and cross-domain generalization (trained in mixed weather but tested in a specific one). This analysis shows whether models can generalize to unseen conditions or require weather-specific training. In all experiments, YOLOv8 consistently achieved superior performance and was therefore selected for detailed weather-specific evaluations.


\begin{figure}[H]
    \centering
    \includegraphics[width=0.34\textwidth]{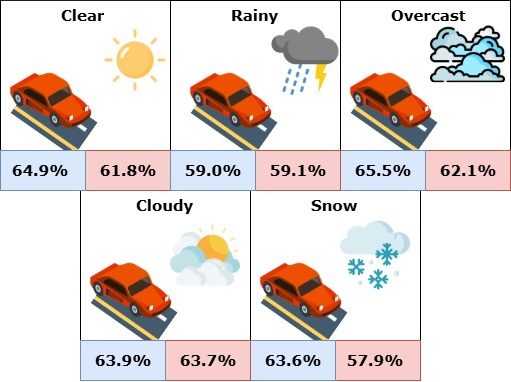}
    \caption{
    Weather‑specific YOLOv8 performance. {In‑Domain (Blue) – trained and tested on the same weather.} {Cross‑Domain (Red) – trained on all weathers, tested on the shown weather}
    }
    \vspace{-0.4cm}
    \label{fig:Weather}
\end{figure}

As shown in Figure~\ref{fig:Weather} in clear weather, in‑domain accuracy reaches 64.9\%, slightly higher performance of the cross domain 61.8\%, indicating a modest improvement. Rainy conditions achieve the lowest overall performance, with same and cross-domain performances showing minimal differences at 59.0\% and 59.1\%, reflecting the challenge posed by rain and the limited benefit of exposure to mixed conditions. Overcast scenes exhibit the strongest gain from domain‑specific training, achieving the highest in-domain accuracy of 65.5\%, while cross-domain accuracy decreases to 62.1\%, suggesting sensitivity to mismatched training conditions. The cloudy weather shows a minimal difference between the performance in domain (63.9\%) and in cross domain (63.7\%), the model generalizes well in this setting. Snow presents the largest performance gap, with in‑domain accuracy at 63.6\% and cross‑domain accuracy at 57.9\%, highlighting poor generalization and the importance of condition specific training to maintain performance.

These results show the importance of weather aware training in CAVs. YOLOv8 performs reasonably in moderate conditions such as cloudy weather, but degrades in extreme conditions such as rain and snow. Aggregating all conditions together is not sufficient, indicating the need for advanced approaches to handle diverse environments.

\section{\textbf{Discussion and Open Research Problems}}
\label{sec:evaluation}

We evaluated the deployment challenges of federated object detection in CAVs, simulating real-world conditions. Our evaluation highlights the need for adaptive aggregation strategies that account for hardware heterogeneity and dynamic conditions, but our study has certain limitations indicated below.   


First, our evaluation was conducted entirely on a simulation-based platform, which, despite incorporating system-level profiling, non-IID data distributions, dynamic client behaviors, and environmental variability, still cannot fully capture the complexity and unpredictability of real-world CAV operations. Second, due to significant time and resource constraints, we were unable to perform additional experiments or multiple trial runs. Third, although three datasets with varied image resolutions and lighting conditions were used to broaden coverage, the scope of this study was restricted to 2D object detection, excluding other critical sensor modalities such as LiDAR, radar, or multimodal fusion. Fourth, we used only baseline aggregation strategies, while we simulated client dropouts, failures, and rejoining to evaluate robustness without interrupting training. Fifth, although communication efficiency is critical in FL, we did not model bandwidth or delays and instead focused on profiling local training resources per client. Sixth, dataset imbalances, particularly varying sample sizes across different weather conditions, limited direct comparability, despite reflecting real-world data imbalance in autonomous driving. Finally, this study did not evaluate security or privacy aspects of FL. 

Despite these limitations, our work highlights key deployment challenges under realistic constraints, focusing on non-IID data, hardware variability, client dynamics, and environmental conditions to establish a foundation for future exploration. 
Based on our comprehensive evaluation, we next identify a set of open research problems from the perspectives of datasets, machine learning models, and aggregation strategy related issues that must be addressed to facilitate real-world FL deployment in CAVs.

\vspace{-0.2cm}


\subsection{\textbf{Datasets}}

Datasets for object detection in CAVs remain limited in their ability to represent real-world complexity. Diversity in environmental conditions is still insufficient; while some datasets capture adverse weather or nighttime driving, others emphasize clear daytime conditions, and combining them into a unified benchmark remains difficult. Benchmarks tailored to federated settings are also absent, even though data in practice is naturally partitioned across vehicles, making systematic comparison across studies challenging. Nearly all datasets are collected once and fixed, offering static snapshots rather than continuous streams of evolving driving environments, with only a few providing limited sequential data. Many rely primarily on 2D bounding boxes, with fewer including 3D annotations or richer contextual labels, and annotation quality often varies across datasets. Annotation practices remain inconsistent, and the absence of standardized dataset design continues to limit reproducibility and deployment readiness. Additionally, there is a lack of standardized data quality metrics to evaluate the usefulness and reliability of existing datasets. In summary, it is essential to curate and collect substantially more high-quality multimodal datasets to enable more effective model training and improved performance. The development of data augmentation techniques, along with systematic approaches for evaluating data quality, also requires further exploration.

\vspace{-0.2cm}

\subsection{\textbf{Object Detection Models}}

Object detection models in CAVs must balance accuracy, computational efficiency, and real-time latency. Despite recent advances, our prior work \cite{11133941} identified challenges in detecting small objects, while the current study further reveals limitations in handling underrepresented classes. More broadly, object detection continues to face significant challenges in handling rare-case long-tail problem, particularly in recognizing unseen, unexpected, or transformed objects in real time. Additionally, most existing models are designed for centralized learning, whereas novel approaches that embed distributed learning such as federated learning remains underexplored. While researchers have begun to interoperate  multimodal data for object detection, there is still a lack of innovative and efficient models that truly leverage multimodal information. Moreover, the robustness and security of models requires deeper investigation, especially in the context of adversarial of malicious attacks, and these concerns are even more pressing while supporting distributed learning. Missing standardized evaluation protocol and benchmarking remains a gap between academic research and regularization and industry need. 

\vspace{-0.4cm}

\subsection{\textbf{Aggregation Strategies Related Issues}}

Federated learning differs from centralized learning by relying on distributed local learning and aggregation. Aggregation plays a critical role in the learning results, and our work demonstrates that aggregation in FL for CAVs must be fault tolerant, capable of handling client dropouts, failures, and rejoining without disrupting training. Although no method (Sync, Async, Semi-Sync) is perfect, future research should focus on adaptive aggregation policies that combine strategic selection with client data relevance, resource awareness, environmental variability, communication constraints, as well as trade-offs between security and privacy. Developing such adaptive approaches will push FL beyond fixed aggregation schemes and support scalable, robust, and deployment-ready solutions for CAVs.

\subsubsection{\textbf{Client Selection Policy}}

In FL for CAVs, client selection is essential for fair and efficient aggregation, as all clients do not contribute equally. Our experiments show that clients with stronger processors and larger datasets provided more effective updates, leading to faster and more stable convergence. However, excessive reliance on them risks dominance of the global model, making the inclusion of resource constrained vehicles essential. The fairness in participation continues to be insufficiently addressed, since diverse clients operating under heterogeneous conditions are vital for robust deployment. Scalability further intensifies this challenge, as selecting fairly and efficiently from an expanding pool of vehicles requires systematic and adaptive policies. Moreover, dataset size alone is not a reliable indicator of utility, as large datasets may lack contextual relevance. Hence, effective client selection requires consideration of data selection to achieve both efficiency and diversity.



\subsubsection{\textbf{Data Selection Policy}}

In FL for CAVs, data selection must balance computational efficiency and learning effectiveness. In our experiments, we observed that even when certain clients dropped out, the global model continued to improve, indicating that their data had already been sufficiently represented. This highlights the difficulty of distinguishing data that contribute meaningfully to the global model from data that offer limited value, which remain unresolved. A further issue is prioritizing data that capture the diversity of driving environments so that rare or underrepresented scenarios are adequately reflected. Equally important is the absence of standardized benchmarks for evaluating data selection strategies, making current approaches difficult to compare and validate. Principled data selection is therefore essential to improve efficiency, maintain balance, and strengthen generalization. Moreover, the interplay between data selection and client selection could further complicate the decision-making process. Jointly optimizing which data samples to use and which clients to involve introduces trade-offs among efficiency, fairness, and model performance.

\subsubsection{\textbf{Resource Aware Policy}}

In FL for CAVs, resource constraints remain a key barrier to practical deployment. Vehicles vary widely in GPU availability, memory capacity, and energy budgets, making uniform training expectations unrealistic. Based on our experiments, tuning parameters such as image resolution and batch size allows both low and high resource clients to contribute without overload. However, this adjustment alone is insufficient, as fluctuating resources still interrupt training and reduce stability. Variability in resource utilization makes it challenging to sustain local training, as energy depletion or memory overload can interrupt participation and disrupt stability. The interaction between training and inference further complicates the problem, since inference must take priority when resources are constrained to ensure real-time decision-making and operational safety. This challenge will be even more severe as autonomous vehicles evolve into computing platforms running multiple concurrent applications, as envisioned in the braoder scope of vehicle computing~\cite{lu2023vehicle}. Therefore, without reliable mechanisms to preserve training progress, interruptions under constrained conditions cause recomputation and inefficiency. Finally, the evaluation of resource-aware FL in CAVs remains inconsistent, as there are no standardized benchmarks to ensure fair comparison and reproducibility across heterogeneous hardware platforms. Establishing such benchmarks, along with designing adaptive mechanisms that dynamically allocate resources while preserving safety-critical inference, is a critical open challenge for the field.



\subsubsection{\textbf{Communication}}

FL reduces communication by transmitting model updates instead of raw sensor data, which is beneficial for CAVs. However, vehicles often operate under unreliable and intermittent network conditions, leading to delayed updates and slower aggregation. Communication challenges are further extended by the need to share bandwidth with other vehicular tasks, where safety-critical messages must always take priority. Intermittent connectivity results in delayed model updates, inconsistent bandwidth, and scheduling conflicts, each of which undermines the timeliness and reliability of model convergence. As the number of participating vehicles increases, these issues become even more severe, raising concerns about congestion and fairness in bandwidth allocation. Despite progress in communication, standardized communication policies that can adapt to fluctuating connectivity while ensuring fairness and preserving safety priorities remain underdeveloped as an open challenge.


\subsubsection{\textbf{Security and Privacy}}

FL reduces communication overhead and avoids raw data sharing, but both security and privacy remain critical barriers to deployment in CAVs. On the security side, model poisoning can corrupt the global model, while backdoor attacks implant hidden triggers that misclassify safety-critical objects. Data integrity and secure aggregation must also be ensured, as client updates are vulnerable to tampering during transmission. On the privacy side, leakage through model updates persists, since adversaries can infer sensitive driving records or reconstruct features despite the absence of raw data. Techniques such as differential privacy and secure aggregation offer protection, but they often introduce significant computational overhead that is difficult to sustain on resource-constrained vehicles. Moreover, new forms of privacy leakage, such as membership inference and property inference, remain insufficiently addressed in federated settings. Finally, beyond developing security enhancement and privacy-preserving techniques, when a security breach is detected or sensitive information must be removed during of after the training process, mechanisms such as federated unlearning need to be explored~\cite{liu2024survey}. However, achieving effective federated unlearning presents significant challenges, including ensuring that removal is both complete and efficient, minimizing the impact on model accuracy, and maintaining consistency across distributed clients. Addressing these challenges is essential for building trustworthy and resilient FL systems in CAVs.





\subsubsection{\textbf{Lighting and Weather}}

Aggregating all available data is insufficient, as our results show that models perform well when training and testing occur under the same weather conditions, but performance declines significantly in cross-domain settings. This underscores the broader challenge of adapting FL aggregation to diverse and dynamic environmental conditions. Weather and illumination introduce domain shifts that degrade generalization, yet current aggregation approaches rarely account for these variations. Rare but safety-critical scenarios such as fog, snow are underrepresented, leaving models vulnerable in adverse conditions. Moreover, vehicles equipped with multimodal sensors provide complementary signals, but effective strategies to integrate these modalities under heterogeneous conditions remain underdeveloped. Finally, the absence of standardized protocols for evaluating aggregation across varying weather and illumination further limits progress, as comparisons remain inconsistent.

\section{\textbf{Conclusion}}
\label{sec:conclusion}

Real-world deployment of FL in CAVs requires balancing detection performance with hardware constraints and environmental heterogeneity. Unlike prior works that emphasize algorithms or accuracy benchmarks, our contribution lies in quantifying the interplay of heterogeneity, resource constraints, and environmental variability, thereby advancing a deployment-oriented perspective for evaluating FL in CAVs. Building on these findings, we also outline future directions and open research challenges to guide scalable, efficient, and reliable real-world deployment of FL in CAVs.

\ifCLASSOPTIONcaptionsoff
  \newpage
\fi





\bibliographystyle{IEEEtran}
\bibliography{IEEEabrv,Bibliography}

\begin{IEEEbiography}
[{\includegraphics[width=1in,height=1.25in,clip,keepaspectratio]{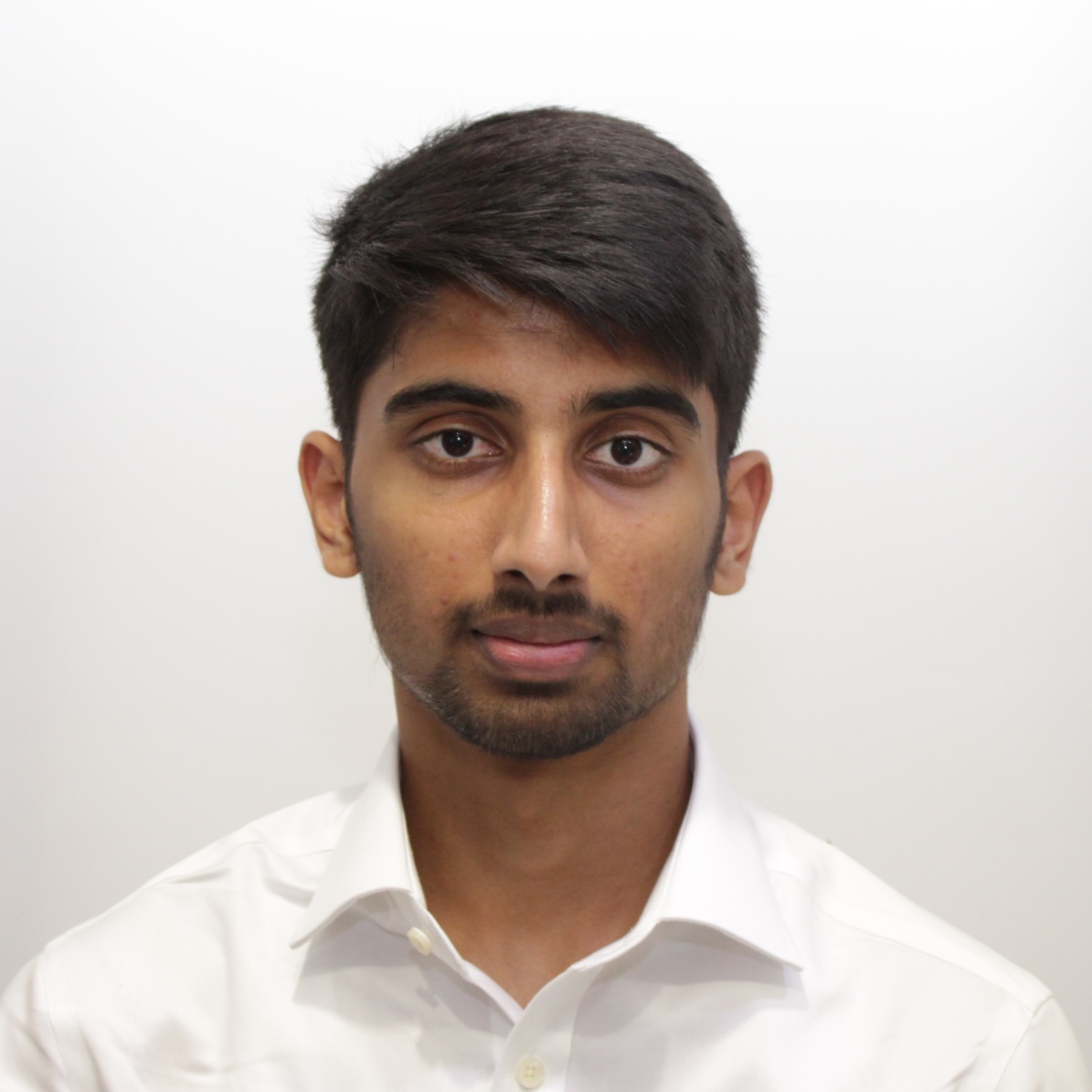}}]{Komala Subramanyam Cherukuri} (Student Member, IEEE) received the B.E. degree in Software Engineering from the University of Greenwich, London, U.K., in 2023, and the M.S. degree in Data Science from the University of North Texas (UNT), Denton, TX, USA. He is currently pursuing the Ph.D. degree in Data Science at UNT. His current research interests include distributed systems, federated learning, and connected autonomous vehicles.
\end{IEEEbiography}

\vspace{-0.5cm}

\begin{IEEEbiography}
[{\includegraphics[width=1in,height=1.25in,clip,keepaspectratio]{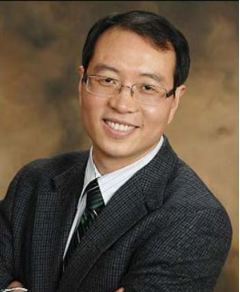}}]{Kewei Sha} (Senior Member, IEEE) received his Ph.D. degree in Computer Science from Wayne State University in 2008. He is currently an Associate Professor of Data Science and Director of Data Science Programs at the University of North Texas. His research interests include the Edge Computing, Security and Privacy, Applied AI, Data Quality, and Intelligent Agent Systems. He serves as an Associate Editor for IEEE IoT Journal, Smart Health, and Multimedia Tools and Applications.
\end{IEEEbiography}

\vspace{-0.5cm}

\begin{IEEEbiography}
[{\includegraphics[width=1in,height=1.25in,clip,keepaspectratio]{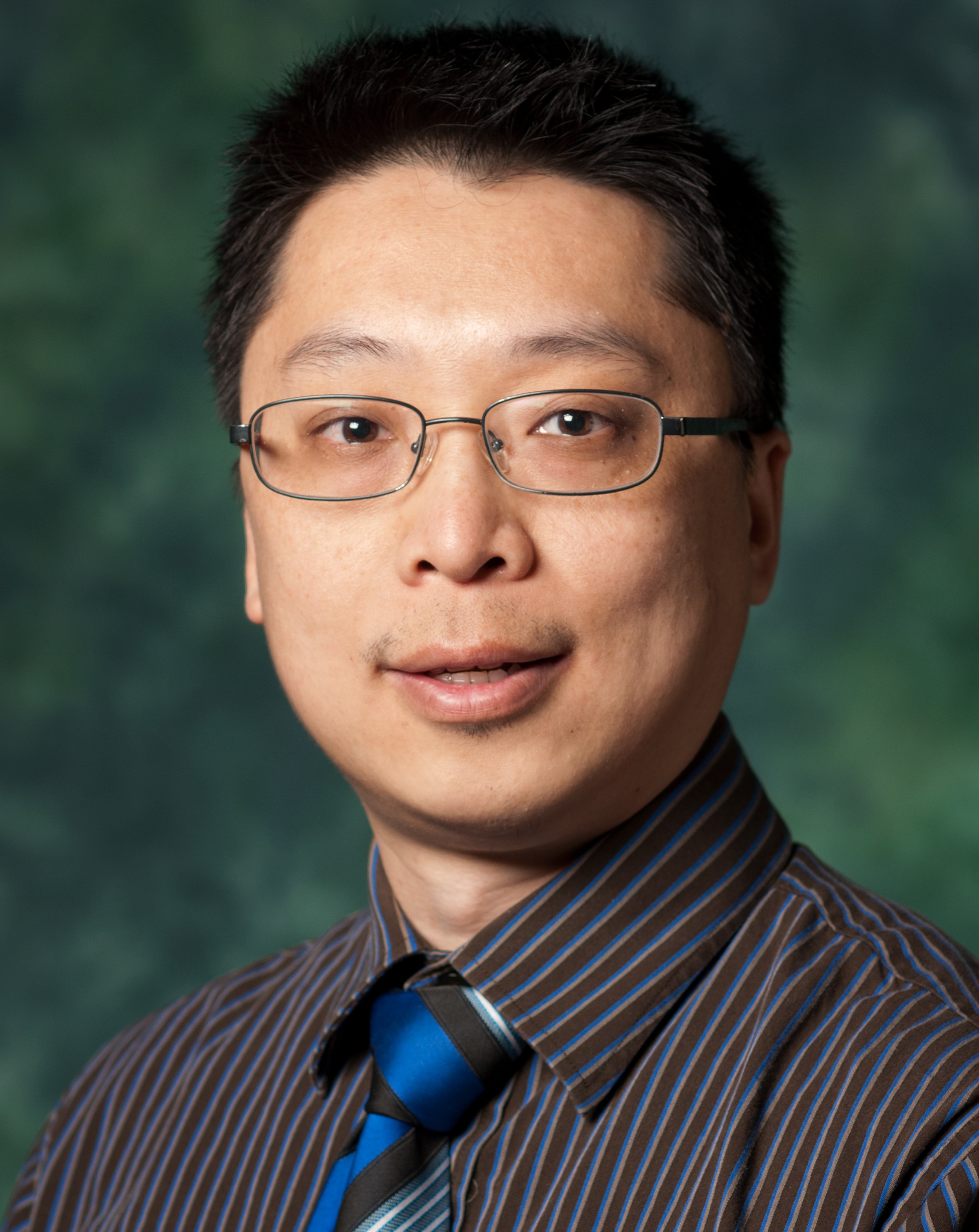}}]{Zhenhua Huang}  is a Professor in the Department of Mechanical Engineering and an Affiliate Professor in the Department of Data Science at the University of North Texas. He received his Ph.D. in Structural Engineering from the University of Illinois at Urbana-Champaign. His research focuses on artificial intelligence and machine learning in civil engineering, hazard mitigation, and transportation infrastructure. He has led and contributed to numerous federally and state-funded projects, published extensively in peer-reviewed journals, and supervised doctoral and master’s students in engineering and data science.
\end{IEEEbiography}

\vfill

\end{document}